%% file: main.tex
\documentclass[letterpaper, 10 pt, conference]{ieeeconf}  
\IEEEoverridecommandlockouts                              
\input{preamble.tex}

\title{\LARGE \bf
Deep Reinforcement Learning with Feedback-based Exploration
}
\author{Jan~Scholten, Daan~Wout, Carlos~Celemin, and Jens~Kober% <-this % stops a space
\thanks{All authors are with Cognitive Robotics Department, Faculty of Mechanical, Maritime and Materials Engineering, Delft University of Technology, The Netherlands and reachable via {\tt\small jan@jjscholten.com}}}% <-this % stops a space

\begin{document}
\maketitle
\thispagestyle{empty}
\pagestyle{empty}

\begin{abstract}
Deep Reinforcement Learning has enabled the control of increasingly complex and high-dimensional problems. However, the need of vast amounts of data before reasonable performance is attained prevents its widespread application. 
We employ binary corrective feedback as a general and intuitive manner to incorporate human intuition and domain knowledge in model-free machine learning. 
The uncertainty in the policy and the corrective feedback is combined directly in the action space as probabilistic conditional exploration. 
As a result, the greatest part of the otherwise ignorant learning process can be avoided. We demonstrate the proposed method, Predictive Probabilistic Merging of Policies (PPMP), in combination with DDPG. 
In experiments on continuous control problems of the OpenAI Gym, we achieve drastic improvements in sample efficiency, final performance, and robustness to erroneous feedback, both for human and synthetic feedback. 
Additionally, we show solutions beyond the demonstrated knowledge.

\end{abstract}
\section{Introduction}
Contemporary control engineering is adopting the data-driven domain where high-dimensional problems of increasing complexity are solved, even if these are intractable from a classic control perspective.
Learning algorithms, in particular Reinforcement Learning \cite{sutton18reinforcement}, already enable innovations in robotic, automotive and logistic applications\cite{pinsler18sample,ferdowsi18robust,bii18ai} and are on the verge of broad application now that data becomes ubiquitous\cite{henderson17matters}.
There are many applications also beyond the classical control engineering domain, such as HIV \cite{ernst07clinical} and cancer treatment schedules\cite{zhao09reinforcement}.
A possibly extension to diabetes treatment could have great impact
% and a possible extension to diabetes treatment 
% --- an extension to diabetes treatment could improve the lives of millions whilst saving expenditure of billions
\cite{dci180007}.
In contrast to model-based control, RL is able to retain optimality even in a varying environment, and modelling of dynamics or control design is not needed.

This study concerns deep RL (DRL), the leading approach for high-dimensional problems that uses neural networks to generalise from observations to actions.
DRL can greatly outperform humans \cite{hessel17rainbow} in virtue of machine precision and reaction time.
However, DRL requires extensive interaction with the problem before achieving final performance.
For real-world systems that have restrictions on interaction, the sample efficiency can be decisive for the feasibility of the intended application \cite{kober13survey}.
Improving sample efficiency is thus essential to the development of DRL and its applications.

In contrast to autonomous learning algorithms, humans are very effective in identifying strategies when faced with new problems.
In many cases we achieve decent performance in the first try, 
despite poorer precision and reaction time that limit final performance.
Indeed, from the sample efficiency perspective, human and RL performance are complementary and incorporating human insight into a learning algorithm is a great way to accelerate it.

\simplefigs{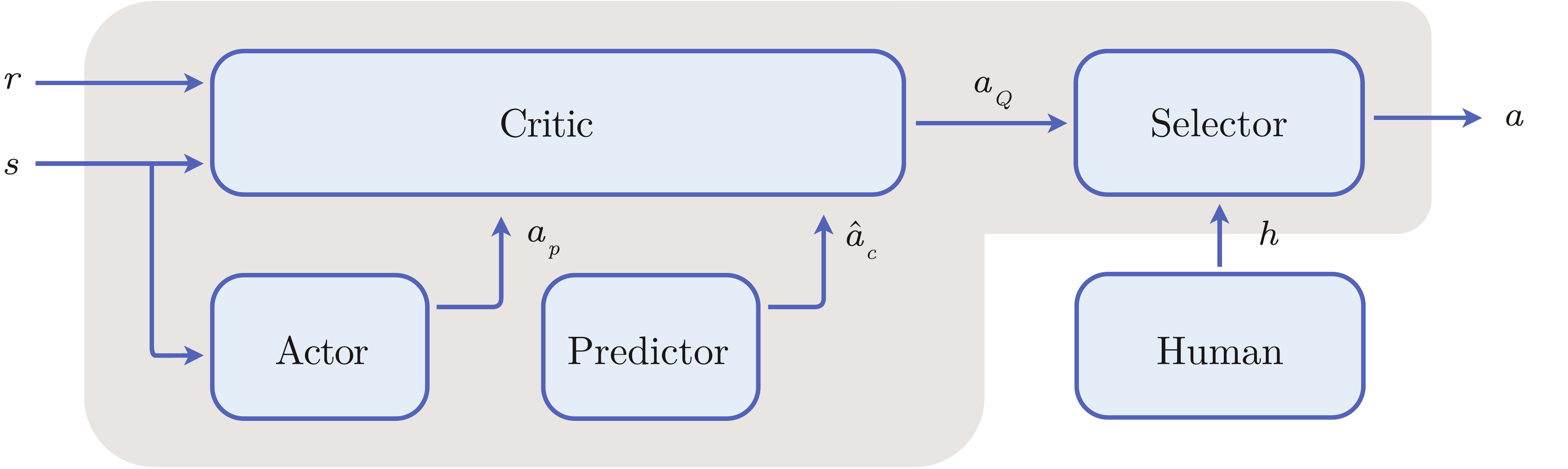}{In the suggested approach, human feedback is combined either with the policy or a prediction of the corrected action. The critic discriminates these with respect to the estimated value of the action. The magnitude of the correction is estimated in the \emph{Selector} and proportional to the estimation variance of the policy. The elements in the grey area constitute an autonomous learner.}{trim=4 0 30 0, clip, width=\linewidth}

Some existing RL methods update the policy with additional human feedback, which is provided occasionally \cite{knox12reinforcement, celemin18fast}. 
In contrast, we propose to keep the original RL process and use the human feedback for exploration, as Nair et al.\ do with demonstrations \cite{nair18overcoming}.
Focussing on DRL, there are methods that learn from a priori demonstrations \cite{vecerik17leveraging} or intermittently collect those (dataset aggregation) \cite{monfort17asynchronous}. In contrast to corrective feedback, demonstrations are not always available or even possible (there may be limitations in the interface or expertise), besides that they may require manual processing \cite{vecerik17leveraging} or simulation \cite{nair18overcoming}.
Likewise, methods that receive preferences between trajectories can be powerful\cite{christiano17deep} but they assume the availability of simulation, which is not generally realistic.
As a general measure, there is evaluative feedback \cite{pinsler18sample} but we believe that there currently is no method that uses (binary) corrective feedback 
\cite{celemin18fast}
to accelerate DRL. Yet for the purpose of conditioning the exploration of continuous control problems this would be a natural choice.
Moreover, corrective feedback promises to be more effective than evaluative rewards especially in larger action spaces \cite{suay11effect}.

\vspace{.5em}
We present a pioneering combination of DRL with corrective human feedback for exploration, to efficiently solve continuous control problems (\fref{fig:ic74}). 
We revisit the question of how the current estimate of the policy is best combined with feedback, and subsequently derive a probabilistic algorithm named Predictive Probabilistic Merging of Policies (PPMP) that improves the state-of-the art in sample efficiency, is robust to erroneous feedback, and feedback efficient. 
Whilst the proposed assumptions remain realistic, the introduced techniques are moreover generic and should apply to many deep actor-critic (off-policy) methods in the field. 

% There are some deep methods that learn from feedback only \cite{warnell17deep,dattari18interactive}
Our approach is motivated by four ideas:
\subsubsection*{Action Selection} 
After first evidence by Knox \& Stone\cite{knox10combining},
it later were Griffith et al.\ \cite{griffith13policy} who made a strong case for how human feedback is given with respect to the action (sequence) and it is most effective to directly adjust the actions when feedback is obtained. 
Their algorithm outperformed other evaluative feedback methods that instead affect the actions indirectly (modification of the policy).

\subsubsection*{Significant Error}
If we consider the early learning phase, where the policy is useless but the human feedback most valuable, we believe feedback is received in case of a significant error as to help the agent develop a notion of the task rather than to communicate refinements.
Indeed, a recent study demonstrated that vigorous initial exploration is beneficial for sample efficiency \cite{haarnoja18soft}. 
Moreover, we argue that the instantaneous precision of human feedback
is then rather coarse (in contrast, corrections for steady-state errors of an almost converged policy may be smaller).
Accordingly, this limitation is quantified by defining a precision $d$ expressing a region of indifference per dimension. 

\subsubsection*{RL for Fine-stage Optimisation}
Reinforcement learning is superior in final performance due to its precision and reaction time \cite{mnih15human}. 
From our point of view, it should therefore be allowed to autonomously optimise during the later learning phases, such that local optima are identified (e.g.\ using gradients) independent from past feedback. 
\subsubsection*{Probabilistic Approach}
Griffith et al.\ \cite{griffith13policy} proposed a probabilistic combination of the policy and feedback distributions to determine the action.
Because the policy and feedback estimates 
are balanced by their respective accuracy, such approaches are very effective and robust.
In the words of Losey \& O'Malley: \emph{`When learning from corrections ... [the agent] should also know what it does not know, and integrate this uncertainty as it makes decisions'} \cite{losey18including}.
We subscribe to this point of view and furthermore emphasise past success of using uncertainty in other fields, such as the Kalman filter or localisation algorithms \cite{thrun05probabilistic}.
However, whereas Losey \& O'Malley estimate the variance in the corrections\cite{losey18including}, we consider the feedback (co)variance fixed ($d$ in \emph{Significant Error}) and argue that the correction size is inversely proportional to the performance of the agent. 

It immediately becomes apparent that some of these ideas align, 
e.g., that corrections are inaccurate (\emph{Significant Error}), but do not need to be accurate, since RL will efficiently identify local optima (\emph{RL for Fine-stage Optimisation}).
However, before connecting the dots, let us complete this motivation with the assumption that, given the assumed area of indifference of \emph{Significant Error} expressed by $d$ (\fref{fig:reconcile}), RL is able to identify the local optimum. In other words, \emph{Significant Error} and \emph{RL for Fine-stage Optimisation} concern overlapping regions and the global optimum is attained if the feedback brings us in proximity.

As a corollary of the above statements, we develop a learning method where actions are obtained by significant modification of the policy in direction of the obtained binary feedback. A probabilistic manner that reflects the current abilities of the agent determines the magnitude of correction.
This method strongly reduces the need for interaction and furthermore improves final performance.
Autonomy and optimality are furthermore preserved, since
there will be no feedback 
when the performance is deemed satisfactory, 
and our method then resorts to its RL principles.

\section{Background}
This study is defined in a sequential decision making context, in which the Markov decision process serves as a mathematical framework by defining the quintuple $(\mathcal{S}, \mathcal{A}, \mathcal{T}, \mathcal{R}, \gamma)$.
This set consists of state-space $\mathcal{S}$, action-space $\mathcal{A}$, transition function $\mathcal{T}: \mathcal{S} \times \mathcal{S} \times \mathcal{A} \mapsto [0,1]$, reward function $\mathcal{R}:\mathcal{S} \times \mathcal{S} \times \mathcal{A} \mapsto \mathbb{R}$, and constant discount rate $\gamma$ \cite{sutton18reinforcement}. 

The computational agent interacts with an environment by taking actions $a_k$ (where convenient, we omit the time index $k$) based on its current state $s_k$ and will then end up in a new state $s_{k+1}$ and receive a reward $r_k$ and human feedback $h_k\in\{-1,0,1\}$ to indicate an advice of the direction in the action space, wherein the agent could explore. 
The objective of the agent is to learn the optimal policy $\pi^*(s)$ that maximises the accumulated discounted reward $R=\sum_i \gamma^ir_i$
And we assume the feedback aligns with this goal. 
Along with the policy contained in a neural network called the \emph{actor}, the agent will have a network called \emph{critic} which learns to predict the \emph{value} of a state-action pair, i.e., the $Q$-function $Q(s, a) = \E{R \given \pi, s_k = s, a_k = a} $. 
This deep actor-critic approach was introduced in \cite{lillicrap15continuous}. 
Their work is the basis of the RL-functionalities used here, such as target networks $\pi'$ and $Q'$ and replay buffers $B$, although we consider our scheme could also be applied to other RL algorithms.

\section{Predictive Probabilistic Merging of Policies}
\simplefigs{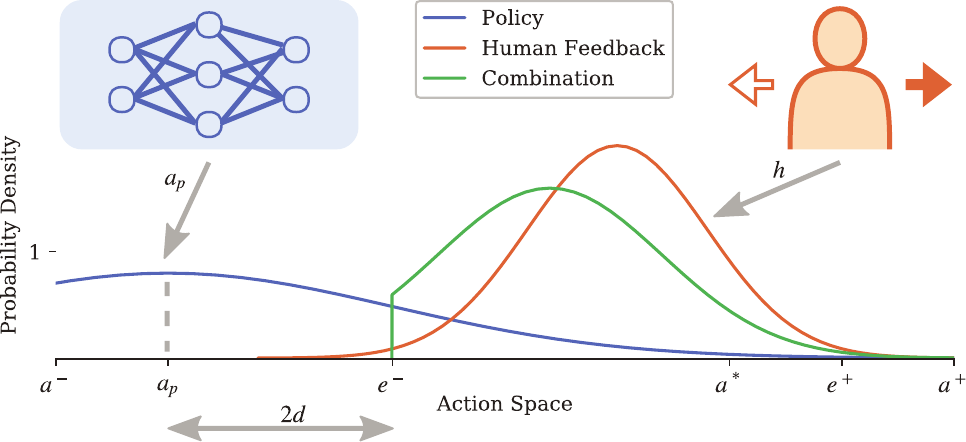}{Using respective covariances, the policy is combined with human feedback in the action space. The resulting distribution on the action that is selected, is truncated such that corrections always have significant effect and the given information cannot dissipate in case of an overconfident policy.}{width=\columnwidth}
\label{sec:design}
The aforementioned point of view materialises in our new learning algorithm PPMP, of which we will discuss each element in one of the following subsections. 

\subsection{Combining Policy Information in the Selector}
\label{sec:combining}

For the sake of this explanation, let us temporally assume non-erroneous corrective feedback $h$ on policy $a_p$ that indicates the relative location of optimal action $a^*$ (all scalar, as in \fref{fig:reconcile}).
With reference to the \emph{Action Selection} statement, our approach is to immediately alter the actors suggested $a_p$ in the direction of $h$, such that the eventually selected action 
$a = a_p +\hat eh$
(the orange distribution in \fref{fig:reconcile}) and $\hat e$ being an estimate of the absolute error $|a_p-a^*|$.

Deriving from the Kalman filter, the unknown magnitude of the error is estimated using the covariance of the policy (the prediction) and the feedback (an observation) in a module that we call the \textit{Selector}.
It is assumed that the magnitude of the error will diminish over time along with the covariance of the policy $\Sigma_{a_pa_p}$. 
With the covariance of the feedback $\Sigma_{hh}$ as a known constant and 
% the covariance of the policy 
$\Sigma_{a_pa_p}$ obtained as described in \sref{sec:multihead}, let
    $\hat e = 
    G\, \text{diag}(c_s) + \mathbf{1}c_o^T$,
 (lines~9-11 in Algorithm~\ref{alg:mine}) where the constant vectors $c$ set the bounds on $\hat e$ as described in the last two paragraphs of this section, 
    $\mathbf{1} = [1, 1, \hdots, 1]^T$ and
    $G = \Sigma_{a_pa_p}(\Sigma_{a_pa_p} + \Sigma_{hh})^{-1}$.
Note that $G \in (0,1)$ (all $\Sigma$ are positive definite by definition) is analogue to the Kalman gain as a dimensionless trade-off measure.
When the policy shows large covariance, the corrections will have larger effect and facilitate vigorous exploration.
And inversely, corrections will be more subtle upon convergence of the policy.
Besides that, the exploration is automatically annealed over time by the decrease of $\Sigma_{a_pa_p}$, its effect is state-dependent and tailored for every action channel, respecting correlations. 

The relation of $G$ to $\hat{e}$ (line~11) is defined using two vectors of which the length equals the dimensionality of the action space.
First, let us discuss the relevance of offset $c_o$.
With reference to \emph{Significant Error}, a lower bound on $e$ is $e^-=a_p+hd$.
Moreover, for the action selection we may further restrict the search by using the fact that $a$ is guaranteed to be closer to $a^*$ than $a_p$ even if 
$\hat e = 2d$. 
Setting $c_o=2d$ accordingly sets an effective lower bound on the corrections.
Apart from the optimisation perspective, it is always desired to apply a significant correction in case feedback is provided.
First, it will avoid frustration of the user, who actually observes the effect of the feedback.
Second, the information is not preserved otherwise. 

The scale $c_s$ allows us to set an upper bound ${e}^+$ for the applied corrections.
From the perspective of using human feedback as exploration, let us consider the case where the policy suggests some negative action and receives $h=1$ since optimality is contained in the positive half of the action space (\fref{fig:reconcile}). 
Although we cannot make any general statements about the reachability of the state-space, it is clear that feedback can only have the intended effect when $\hat{e}$ is large, else there is no escape from the wrong half of the action space. 

\subsection{Integrating the Selector with Autonomous Learning}
The ideas established in the previous paragraph raise requirements for the eventual algorithm. 
The probabilistic combination of the policy and the feedback results in off-policy data in a continuous action space.
As critic-only methods are suitable for a discrete action space whilst actor-only methods are on-policy, an off-policy actor-critic scheme remains as the evident choice.

\fref{fig:ic74} illustrates how the system is interconnected and the actions selected. 
It is assumed that the human provides binary feedback signals $h$ occasionally and bases this on the observed state sequence (and possibly the actions).
Delays between human perception and feedback are not taken into account. 
In order to memorise and generalise the advised corrected samples, those corrected actions are estimated in the \emph{predictor}, a supervised learner further discussed in \sref{sec:predictor}.
First, the $Q$-filter (critic) decides whether the policy's action $a_p$ or the estimated corrected action $\hat{a}_c$ is preferred as the suggested action $a_Q$ (line~7). 
Then, in accordance with the description in the previous paragraph, the selector module adjusts $a_Q$ with respect to $h$ and we arrive at the actually executed action $a$ (line~11).
In case feedback is not provided the algorithm relies on its own policy, including exploration noise. Autonomy is hereby preserved. 

\subsection{Multihead Actor Network}
\label{sec:multihead}
In contrast to DDPG \cite{lillicrap15continuous} we need not only to estimate an action, but furthermore to estimate the covariance in this estimate. 
In \cite{gal16uncertainty} is established that the uncertainty over a deep neural networks output may be obtained from multiple passes with dropout.
However, in the context of RL, \cite{osband16deep} reports how a multihead neural network that maintains multiple hypotheses is a more consistent approach to generate posterior samples than dropout.
Whereas in their study the eventual purpose of the multihead network is to use the posterior for exploration rather than to quantify confidence (as desired for our approach), 
\cite{rupprecht17learning} indeed establishes how the multiple hypotheses provide for accurate estimation of abilities in a deep learning classification problem.
As it is furthermore desired to have efficient and scalable estimation, we apply the multihead architecture as discussed in \cite{osband16deep} to the actor network.

Effectively, the modification of a regular actor network to its multihead counterpart results in $K$ copies of the output layer that estimate the optimal action $a_j=\pi_j(s\, \given \psi)$ (line~5), where $j$ indicates the head and $\psi$ is the parameter set of the network. 
For the training, we establish an extension to the sampled policy gradient in \cite{lillicrap15continuous} that features individual values of $\nabla_a Q$ and $\nabla_\psi \pi$ for each head.
This sampled multihead policy gradient is given by
\begin{equation}
    \nabla_{\psi} J^{\pi} \approx
            \frac{1}{N}\sum^N_{i=1}
                \left.\nabla_aQ(s,a|\theta)\right\rvert_{a=\pi(s_i)}
                \left. \nabla_{\psi}\pi(s|\psi)\right\rvert_{s=s_i},
\label{eq:spg_mh}
\end{equation}
with a slight abuse of notation in the row-wise expansion of $\nabla_aQ$ 
that contains evaluations for all $K$ policies in $\pi$.
To determine the policy during action selection,
we choose to randomly select a head $j_e$ per episode, preserving both temporal consistency and compliance with multimodalities (not preserved when averaging).
For the training of the critic (line~13 in Algorithm~1),
$\pi_{j_i}'(s_{i+1}\given \psi')$ is evaluated for $j_i$, the same head as in $a_i$.

\begin{algorithm}[t]
%   \caption{$K$-Head PPMP}
  \caption{Predictive Probabilistic Merging of Policies}
  \label{alg:mine}
  \begin{algorithmic}[1]
    \Initialize{\strut Neural network parameters $\theta, \theta', \psi, \psi', \phi$\\
    Replay buffers $B$ and $B_c$\\
    % Feedback cov. $\Sigma_{hh}$ scale $c_s$, and offset $c_o$}
    Feedback covariance $\Sigma_{hh}$  \\
    Scale $c_s$, and offset $c_o$}
    \vspace{.35em}
    \For{episode $e = 1$ to $M$}
      \Initialize{Ornstein-Uhlenbeck process $\nu$ \\ Randomly set active head $j_e$}
    \vspace{.2em}
      \For{timestep $k$ = 1 to $T$}
        \State $a_p \gets \pi_j(s_k\given \psi) + \nu_k$
        \State $\hat{a}_c \gets P(s \given \phi) + \mathcal{N} (0,\sigma_a)$
        \State $a_Q \gets \argmax_a Q(s,a)\rvert_{s = s_k, a=a_p \lor a=\hat{a}_c}$
        % \State Obtain feedback $h_k$
        \If{Feedback $h_k$ is given}
            \State $\Sigma_{a_pa_p} \gets$ cov$\left(\pi(s_k\given \psi)\right)$
            \State $ G\gets \Sigma_{a_pa_p}(\Sigma_{a_pa_p} + \Sigma_{hh})^{-1}$
            \State $a_k \gets a_Q + (G\, \text{diag}(c_s) + \mathbf{1}c_o^T)h$
            \State Store $(s_k, a_k)$ in $B_c$
        \EndIf
        \State Obtain $s_{k+1}$ and $r_k$ by executing $a_k$
        \State Store transition ($s_k, a_k, r_k, s_{k+1}, j_{e}$) in $B$
        \State Sample $N$ tuples $(s_i, a_i, r_i, s_{i+1}, j_i)$ from $B$
        \State Compute target $Q$-values~~~~~~~~~~~~~~~~~~~~~~\linebreak
        \hspace*{1.4cm}
        $y_i = r_i + \gamma Q'\left(s_{i+1},\pi_{j_i}'(s_{i+1}\given \psi' )\given[\big] \theta'\right)$
        \State Update $\theta$, $J^Q = 
        \frac{1}{N}\sum^N_{i=1}\left(y_i - Q(s_i,a_i\given\theta)\right)^2$
        \State Update $\psi$ using multihead policy gradient \eref{eq:spg_mh}
        \State Randomly sample $N$ transitions $(s_i, a_i)$ from $B_c$
        \State Update $\phi$, $J^P = 
        \frac{1}{N}\sum^N_{i=1}\left(P(s_i\given \phi) - a_i)\right)^2$  
        \State Update target network $Q: \, \theta' \gets \tau \theta + (1-\tau)\theta'$
        \State Update target network $\pi: \, \, \psi' \gets \tau \psi + (1-\tau)\psi'$
      \EndFor
    \EndFor
  \end{algorithmic}
\end{algorithm}

\subsection{Predictor Module}
\newcommand{\ach}{\ensuremath{\hat{a}_c}}
\label{sec:predictor}

The corrected actions are estimated as $\ach{} = P(s\given \phi)$, where $P$ is the prediction network with parametrisation $\phi$, trained with human-corrected samples $(s, a)$ from buffer $B_c$.
Whilst these predictions can greatly improve the performance especially during the early learning stage (where the improvements need to take place), taking the predicted actions has two important disadvantages. 
First, the corrections and their estimates are coarse improvements that primarily aim to explore. 
The eventual performance is limited and at some point the actor will perform better and the predictor's influence needs to be scheduled away. 

A second problem is that the predictor generalises from few feedback samples and its policy may not be very expressive. 
As a corollary, the variance in the interactions is reduced and this impedes the learning from this data.
As clearly demonstrated in \cite{de15importance}, learning from data generated by a stationary policy will cause for instability, presumably because of overfitting. 
In addition, we suspect that the \verb|Adam| optimiser \cite{kingma14adam} may become over-confident in its gradient estimate (which is now artificially consistent) and raises at least some of the adaptive learning rates to an unstable value.
In \cite{de15importance} the problems are overcome by collecting data with a random or pristine policy. 
Accordingly, we disable the predictor during the first $N_p$ non-corrected samples.
As a second countermeasure, we inject noise to the estimates with variance $\sigma_{\ach{}}$ (line~6), such that the original distribution is somewhat restored and the variance problems partly alleviated. 
Finally, note that in successful actor-critic learning, the critic learns faster than the actor \cite{konda00ac}. %, grondman12ac}. 
We can therefore interleave $\hat{a}_c$ and $a_p$ using a $Q$-filter that selects the action with the greatest value (line~7).
Besides the retaining of buffer variance, emphasis will now be scheduled towards the actor upon its convergence so the $Q$-filter also solves the first problem (transcending the predictor performance).
Because the critic needs to be learned before it can correctly schedule, it is enabled after $N_Q$ samples. 
Note that, in contrast to the use of direct scheduling heuristics \cite{griffith13policy}, there is a wide range in which $N_Q$ and $N_P$ are successfully set (\fref{fig:additive}).

\section{Implementation and Evaluation}
Our code is available at \href{https://github.com/janscholten/ppmp}{github.com/janscholten/ppmp}.
All five neural networks (2 for the actor, 2 for the critic, and one for the predictor) are of size (400, 300) and use ReLU activation (except for hyperbolic tangent output layers that delimit actions within their bounds). We train with \verb|Adam| \cite{kingma14adam}, using learning rates of 0.002 for the critic 0.005, 0.0001 for the actor and 0.0002 for the predictor. The actor has $K=10$ heads. 
The soft target mixing factor $\tau = 0.003$.
The initial variance in the network initialisations is 0.001.
The buffers $B$ and $B_c$ have size 1M and 1600 respectively and the minibatch size is 64. 
The discount rate is $\gamma=0.99$. The OU-process has volatility 0.3, damping 0.15 and timestep 0.01.
The selector and predictor have (as a fraction of the action range per channel) resolution $d=0.125$, scale $c_s=0.5$ and variance $\sigma_{\hat{a}_c} = 0.025$.
The correction variance is set to $\Sigma_{hh}=1\cdot10^{-8}$
The predictor and $Q$-filter are enabled after $N_P=1500$ and $N_Q=4000$ samples respectively.

\simplefig{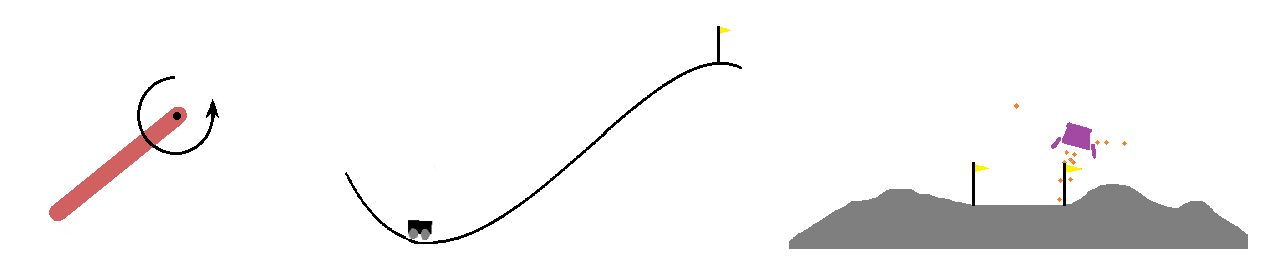}{From left to right: Pendulum-v0, MountaincarContinuous-v0 and LunarLanderContinuous-v2 from the OpenAI gym \cite{brockman16gym}. The respective goals in these underactuated problems is to swingup and balance, drive up the mountain and gently land between the flags.\vspace{-2mm}}
For benchmarking purposes we regard the problem set in \fref{fig:envs}.
The continuous state space of the pendulum environment consists of x- and y-positions of the tip and the angular velocity. 
The control input is the applied torque. Negative reward is given both for displacement from the upright equilibrium and for applied torque.
An episode lasts 200 timesteps.
The mountain car's state space consists of position and speed and it is controlled by a (reverse) driving force (again, all continuous). 
When the car reaches the flag, the episode is over and a reward of 100 is obtained, discounted by the cumulative squared action.
The state space of the lunar lander has eight dimensions, both continuous (positions/velocities) and binary (leg contact). 
It is controlled with two inputs, one for the main engine and a second for the steering rockets.
An episode ends upon soft landing to rest (100 points) or crashing (-100), and navigation yields between 100 and 140 points. 
The applied actions are discounted. 
Unsolved episodes terminate after 1000 timesteps. 
%Each algorithm is tested ten times with the same set of random seeds (from another generator), such that the environments feature equal stochasticity for the different implementations. 

To account for inconsistencies in human feedback \cite{griffith13policy}, we use synthesised feedback (oracle). 
To study applicability, we additionally test with human participants.
The oracle compares $a$ with a converged policy. 
We apply the assumed distance $d$ as a threshold for the feedback, but not for the \verb|DCOACH| implementation (the performance would be unfairly hindered by an assumption that the authors do not make in their work).
The feedback rate is controlled with a biased coin-flip and annealed over time.
To infer robustness we apply erroneous feedback, implemented as in \cite{celemin18fast}.
% also considered and

The results with human participants were obtained from three participants in the age of 20 to 30 with different backgrounds.
For each algorithm, they were given a single demonstration and a test run possibility to get familiar with the interface. 
The subsequent four runs were recorded. 

We evaluate PPMP and its ablation PMP (without the predictor) and compare with DDPG \cite{lillicrap15continuous} and DCOACH
\cite{dattari18interactive}
% (a non-RL deep method that learns from corrective feedback only)%
% \footnote{
% Implementations are from 
% {\color[rgb]{0.2,0.31,0.49}\href{https://github.com/pemami4911/deep-rl}{P.~Emami}} and 
% {\color[rgb]{0.2,0.31,0.49}\href{https://github.com/rperezdattari/Interactive-Learning-with-Corrective-Feedback-for-Policies-based-on-Deep-Neural-Networks}{R~P\'{e}rez Dattari}} on Github.com.}.
(a non-RL deep method that learns from corrective feedback only).
Implementations are from 
{\color[rgb]{0.2,0.31,0.49}\href{https://github.com/pemami4911/deep-rl}{P.~Emami}} and 
{\color[rgb]{0.2,0.31,0.49}\href{https://github.com/rperezdattari/Interactive-Learning-with-Corrective-Feedback-for-Policies-based-on-Deep-Neural-Networks}{R~P\'{e}rez Dattari}} on Github.com.
We generated ten random seeds (with another random generator) which we applied to ten runs of each algorithm respectively, such that the environments feature equal stochasticity for the different implementations. 
We evaluate the results on four criteria: sample efficiency, 
feedback efficiency, final performance, and robustness to erroneous feedback.

\section{Results}
\simplefigs{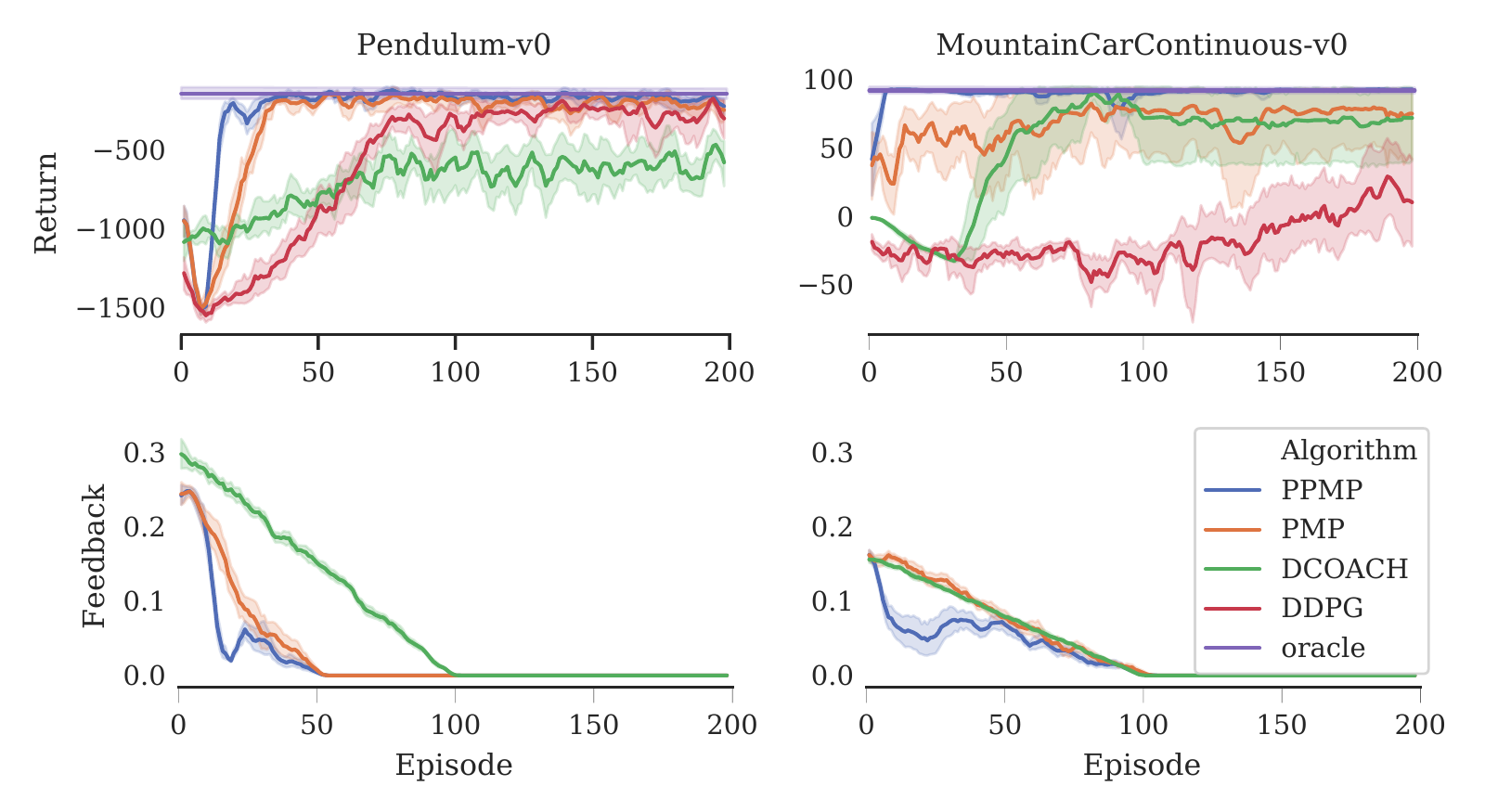}{Our methods PPMP and its ablation PMP (without prediction) outperform all baselines. Depicted is the moving average of ten evaluations (window size 5) along with the feedback rate.\vspace{0mm}}{trim=10 9 12 10, clip, width=1\columnwidth}

\fref{fig:perf} shows that our methods outperform other algorithms in every respect, and whereas the baselines seem to suit one problem in particular, PPMP is consistent. 
DDPG agents only learn good policies within the range of 200 episodes in the Pendulum problem, for the other environments its poor sample efficiency is even more outspoken.
From a comparison with PMP, we infer that the predictor module provides for better sample efficiency, more consistent performance, greater final performance and a reduction of the feedback demand. 
\fref{fig:error} shows the effect of erroneous feedback.
In virtue of value-based learning, our methods prove very robust to erroneous feedback and there is no serious impediment of final performance.
In contrast, DCOACH is greatly affected and fails for error rates beyond 10\%. 
\simplefigs{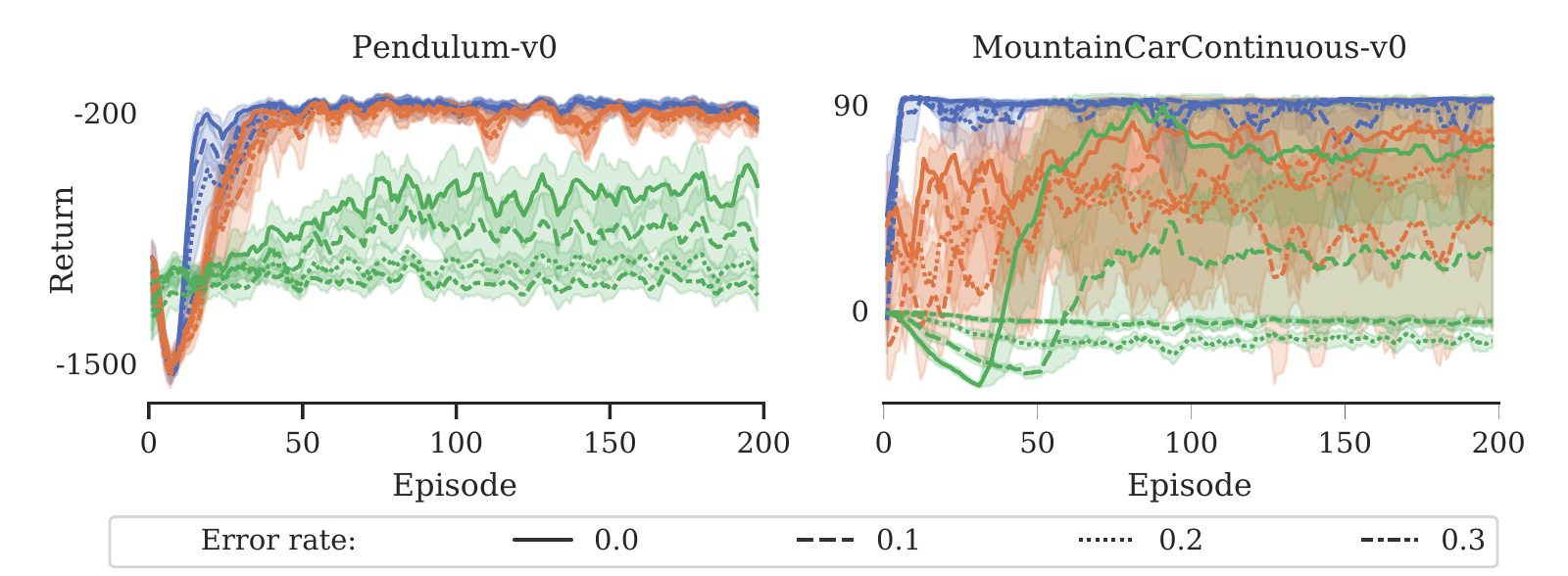}{In case of erroneous feedback, our method proves to be robust and the sample efficiency is hardly affected. The curves with perfect feedback are equal to those in \fref{fig:perf} and the same legend applies.\vspace{-2mm}}{trim=10 2 13 8, clip, width=1\columnwidth}
\simplefigs{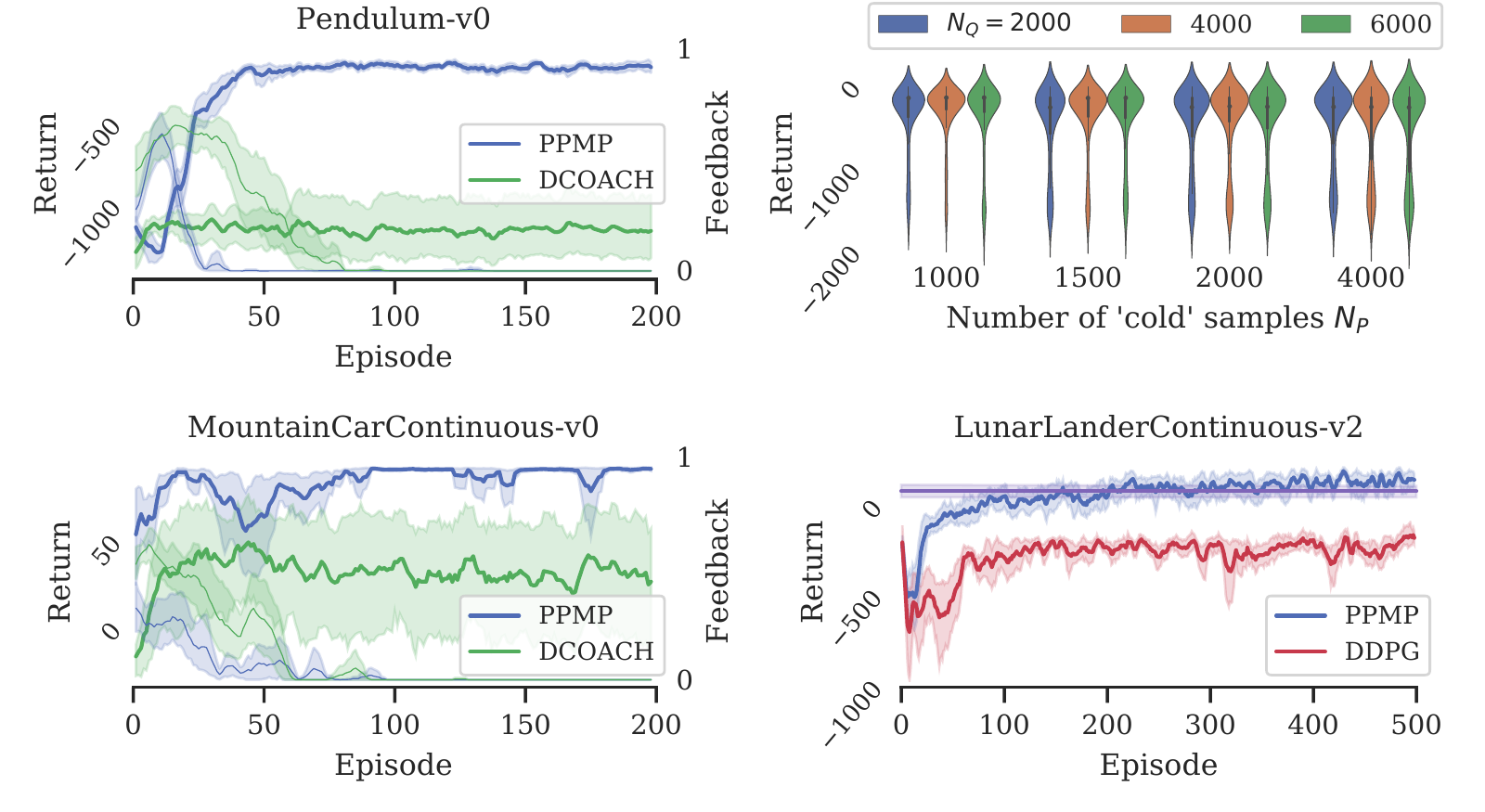}{\textbf{Left:} Performance with human participants, averaged over 12 experiments (feedback rate is the thinner line). \textbf{Top right:} A sensitivity analysis of the introduced hyperparameters. \textbf{Bottom right:} PPMP learns to land, and thereby outperforms the oracle that only knows how to fly.\vspace{-2mm}}{trim=10 9 13 1, clip, width=1\columnwidth}

In \fref{fig:additive} additional results are presented. 
On the left, results from human participants confirm our findings from simulated feedback.
DCOACH obtains a considerable amount of feedback, but inconsistent feedback causes failure nonetheless.
PPMP is more feedback efficient, learns fast, and consistently attains great final performance. 
There are some set-backs in performance in the Mountain Car problem, presumably a result of participants that assume the learning is finished after some early success.

Top right in \fref{fig:additive} is a sensitivity analysis for the new hyperparameters $N_Q$ and $N_p$. 
We compare the distribution of the return during the first 15000 timesteps in the Pendulum domain.
As stated in \sref{sec:predictor} the new parameters neither require meticulous tuning nor cause brittleness. 

Next, in the bottom right, let us consider a typical use case where the feedback has limited performance and is not able to fully solve the problem itself. 
This scenario is different from the previous studies (with erroneous feedback), where incidental mistakes may have been overcome by low-pass dynamics and generalisation but corrections eventually extended to the end-goal. 
We use the Lunar Lander environment, where the oracle is now partial as it knows how to fly but can not land. 
The sequel of the problem is thus left to the agent. 
It is emphasised that the reward function of this environment stresses stable operation by assigning great negative reward to crashes. 
Only the last 100 reward units that our method obtains correspond to having learned to land properly. 
As such, our method allows to solve a problem that is otherwise not feasible. 

\section{Conclusion}
%A disadvantage of Deep Reinforcement Learning is the excessive need for data (thus interaction), which limits applicability to real-world scenarios. 
This work discusses how binary corrective feedback may be used as a probabilistic exploration signal in DRL in order to improve its sample efficiency. 
By slight modification of an off-policy algorithm (here DDPG) the uncertainty in the policy was obtained and coupled with the magnitude of the correction induced by the feedback. 
To generalise corrections and improve memorisation, a predictor network provides estimates of the corrected policy, which can substitute for the actual policy when it increases the value of the state-action pair (especially during early learning).
Our method, Predictive Probabilistic Merging of Policies (PPMP), is easily implemented and makes realistic assumptions about the feedback: it does not need to be a full demonstration, expertise may be limited, we do not assume feedback is abundant, neither do we require simulation or post-processing. 
Nevertheless, PPMP consistently improves on sample efficiency, final performance and robustness in comparison to pure RL (DDPG) and learning from corrections only (DCOACH), both for simulated and human feedback. 

\addtolength{\textheight}{1cm}   % This command serves to balance the column lengths on the last page of the document.
\vspace{.5em}
A first topic further research should address, is how PPMP carries over to real-world scenarios. 
Although the scalability and robustness are promising, the applicability is not yet proven by this work. 
From the possible extensions to this study, an exciting avenue would be to refine the probabilistic part.
In particular, we would like to dispose of the Gaussian approximation and connect with full distributional learning \cite{bellemare17distributional, barth-maron18distributed}, possibly combined with uncertainty-handling estimation of human feedback \cite{wout19learning}. 
This could give better estimates of the abilities and allow for more sophisticated action selection, e.g., posterior or Thompson sampling\cite{osband16deep}.
% \section{Acknowledgement}
% This could not have been without coffee from grant 561.
%%%%%%%%%%%%%%%%%%%%%%%%%%%%%%%%%%%%%%%%%%%%%%%%%%%%%%%%%%%%%%%%%%%%%%%%%%%%%%%%
\bibliographystyle{ieeetr}
\bibliography{references}%
\end{document}

%% file: preamble.tex
\usepackage{cite}
\usepackage{amsmath}
\usepackage{amssymb}
\usepackage{mathtools}
\usepackage{algorithm}
\usepackage{graphicx}
\usepackage[]{algpseudocode}
\usepackage{varwidth}
\usepackage{blindtext}
\usepackage{color} % drafting only
\makeatletter
\let\NAT@parse\undefined
\makeatother
\usepackage[hidelinks]{hyperref} 

\mathtoolsset{showonlyrefs} % Turn off numbering on unused equations

\newcommand{\eref}[1]{\eqref{#1}}
\newcommand{\fref}[1]{Fig.~\ref{#1}}
\newcommand{\sref}[1]{Sec.~\ref{#1}}

\newcommand{\simplefig}[2]{	
	\begin{figure}%
	\centering
	\includegraphics[width=\columnwidth]{#1}%
	\caption{#2}%
	\label{fig:#1}%
	\end{figure}}
	
\newcommand{\simplefigs}[3]{	
	\begin{figure}%
	\includegraphics[#3]{#1}%
	\caption{#2}%
	\label{fig:#1}%
	\end{figure}}
	
\algnewcommand{\Inputs}[1]{%
  \State \textbf{Inputs:}
  \Statex \hspace*{\algorithmicindent}\parbox[t]{.8\linewidth}{\raggedright #1}
}
\algnewcommand{\Initialize}[1]{%
  \State \textbf{Initialize:}
  \Statex \hspace*{\algorithmicindent}\parbox[t]{.8\linewidth}{\raggedright #1}
}

\DeclareMathOperator{\evsym}{\mathbb{E}}
\newcommand\E[1]{\ensuremath{\evsym \!\left[#1\right]}}

\DeclareMathOperator*{\argmax}{arg\,max}

\newcommand\givenbase[1][]{\;#1\lvert\;}
\let\given\givenbase

\DeclarePairedDelimiterX\Basics[1](){\let\given\sgiven #1}
\graphicspath{{./graphics/}}